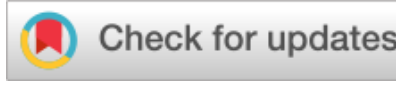

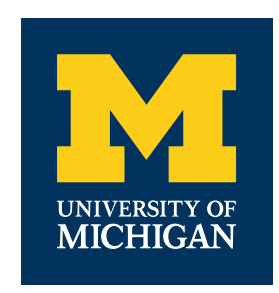

# Automatic Conversion of NICE Guidelines to an Executable Computational Model Using Large Language Models

Ashvin Gupta[1] | Denys Prociuk[2] | Alessandra Russo[1] | Brendan C. Delaney[2]

[1]Department of Computing, Imperial College London, London, UK | [2]Department of Surgery and Cancer and I-X/Digital Foundry, Imperial College London, London, UK

**Correspondence:** Ashvin Gupta (ag619@ic.ac.uk)

## ABSTRACT

**Introduction:** The UK National Institute for Health and Care Excellence (NICE) produce guidelines that provide evidence-based recommendations to support clinical care across England and Wales, but remain available in unstructured natural language form. Converting these guidelines into computable, logically coherent representations is an active area of research yet existing approaches typically focus on individual diseases, require substantial manual encoding, and do not scale. Recent advances in large language models offer an opportunity to automate much of this translation process.

**Methods:** We present an end-to-end approach that automatically converts textual clinical guidelines into an executable model capable of generating explainable patient-specific recommendations. Our approach uses a stepwise LLM-based transformation with in-context examples that can be customized to the guideline of your choice. Each step generates human-inspectable intermediate artifacts, ensuring full transparency and modifiability. We apply the approach to both pancreatic and lung cancer NICE guidelines and use expert human review to assess the alignment of the produced rules as well as evaluating the executable model over 20 pancreatic cancer patient vignettes.

**Results:** Human experts review demonstrated strong alignment between the natural language guidelines and the generated executable models, with the majority of guideline recommendations translated correctly. Most discrepancies involved partial omissions of specific details rather than incorrect logic, and instances of hallucinated or fundamentally incorrect rules were rare. When executed on the vignettes, the resulting executable models produced patient-specific recommendations with an $F1$ score of 82.5%.

**Conclusion:** This work demonstrates that LLMs can be used to automatically transform natural language NICE guidelines into interpretable and executable models. The models preserve guideline structure, allow transparent inspection and modification, and can be executed to generate patient-specific recommendations. Our findings highlight the feasibility of automated guideline generation, opening the door to scalable computable guidelines.

## 1 | Introduction

Primary care is typically the first point of contact for patients presenting with early-stage health concerns where clinicians often assess nonspecific symptoms and determine appropriate diagnostic actions. Diagnostic errors, defined as failures to establish an accurate and timely diagnosis, occur in up to 20% of cases and are particularly consequential in progressive conditions such as cancer [1]. Such errors lead to treatment delays, inferior outcomes, and increased costs [2].

To support clinical decision-making and mitigate diagnostic error, the National Institute for Health and Care Excellence (NICE) publishes evidence-based diagnostic and management guidelines across a wide range of conditions in the UK. These guidelines encode clinical reasoning largely through conditional "if–then" statements expressed in natural language (NL), rather than in a machine-executable form. As a result, their direct use in clinical decision support systems remains limited. Consequently, formalizing guidelines into computable representations has been a decades-long research focus, yielding frameworks such as PROforma [3], Arden Syntax [4], and GLIF [5] which all require extensive manual knowledge engineering.

As reviewed by Scott et al. [6], computable guideline formalisms should exhibit high fidelity (faithfulness to the source guideline), traceability (clear correspondence between narrative text and formal logic), and consistency (the ability to be systematically applied across guidelines). To date, there are no widely adopted methods for the automatic conversion of free-text clinical guidelines into executable, logically structured models. Scott et al. [6] further argue that the development of an intermediate logical representation is a critical step toward scalable computable guideline formalization.

Answer set programming (ASP) is a declarative programming paradigm well suited to knowledge representation and reasoning under complex, conditional logic. The ASP formalism [7] consists of a set of constants and a set of relations (predicates) over those constants, whereas an ASP program is an executable model consisting of a collection of logical rules expressed in such a formalism. Rules are typically written in the form: $h$ :- $b_1$, ..., $b_n$ where $b_1$, ..., $b_n$ (referred to as the *body* of the rule) collectively specifies the conditions upon which the *head* h can be inferred. Such rules can be read as: if all conditions in the body hold, then the head holds. For example, the rule *offer(*"pancreatic protocol CT") :- *have*("obstructive jaundice") states that a pancreatic protocol CT scan should be offered if obstructive jaundice is present. Rules with an empty head, that is, :- $b_1$, ..., $b_n$, represent constraints. They state that the conditions $b_1$, ..., $b_n$ should not hold together. In addition to rules with deterministic conclusions and constraints, the ASP formalism also supports rules that include *choice constructs* in the head or the body of the rules. Rules with choice constructs in the head, allow multiple alternative conclusions to be inferred, while choice constructs in the body allow rules to fire when a specified number of alternative conditions are satisfied. Importantly, executable models expressed in ASP are inherently transparent as every conclusion can be traced back to the specific rules and conditions that support it. These qualities make ASP particularly well aligned with the explainability requirements in this setting. For a formal introduction to ASP, see Vladimir Lifschitz [8].

Recent advances in large language models (LLMs) offer new opportunities to automate the translation of unstructured clinical text into structured representations. LLMs have demonstrated strong performance in extracting medical entities, summarizing clinical narratives, and supporting decision reasoning [9, 10]. Several studies have explored the use of LLMs for translating NL into logic programs. Pan et al. [11] propose an iterative refinement framework using solver feedback, while Coppolillo et al. [12] fine-tune language models on NL-to-ASP pairs. Kalyanpur et al. [13] introduce a critic-based approach that revises generated code based on execution outcomes. However, these methods largely focus on simplified or synthetic inputs and do not address the linguistic complexity, conditional dependencies, and domain-specific semantics found in real clinical guidelines.

The most appropriate related work is that of Ishay et al. [14], who propose a modular approach that decomposes the translation task into constant extraction, predicate identification, and rule generation using in-context learning. This structured approach is particularly well suited to clinical text, as it enables inspection and correction at each intermediate stage. Our work builds directly on this paradigm, extending it to NICE diagnostic guidelines and evaluating both logical fidelity and performance in clinical recommendations (referred to also as *downstream performance*).

## 2 | Methods

In this paper, we present a two-stage framework (see Figure 1): Data-to-knowledge (D2K) and knowledge-to-performance (K2P), where the "data" consists of clinical guideline text. In D2K, the framework uses LLM-based prompting to automatically convert NICE diagnostic guidelines into an executable model expressed in ASP through a transparent, stepwise automated process. In K2P, the framework takes as input patient-specific features and the generated executable model and uses a reasoning engine [15] to produce guideline-conformant recommendations. This approach corresponds to Level 4 (executable) computable biomedical knowledge as defined by Boxwala et al. [16].

### 2.1 | Data

We consider the NICE guidelines "Pancreatic Cancer in Adults: Diagnosis and Management (NG85)" [17], "Suspected Cancer: Recognition and Referral (NG12)" [18], and "Lung Cancer: Diagnosis and Management (NG122)" [19]. For both guidelines, we focus on sections relevant to diagnosis and staging and organize them into two separate datasets.

To evaluate the clinical performance of the generated executable models, a team of oncologists synthesized 20 pancreatic cancer patient vignettes. Each vignette describes patient demographics (e.g., age, sex), presenting symptoms, imaging findings, pathological results, genetic markers, and treatment decisions, all expressed in unstructured NL.

### 2.2 | Data to Knowledge

As illustrated in Figure 1, the D2K stage consists of a three-step process: constant extraction, predicate generation, and rule generation. Each step requires a designed prompt containing general information about the task followed by six in-context examples, an example of which can be seen in Figures 2–4. The same examples are used across the three stages to maintain semantic coherence between outputs.

The design of the prompts was informed by two key principles. First, the in-context examples were selected to cover the full

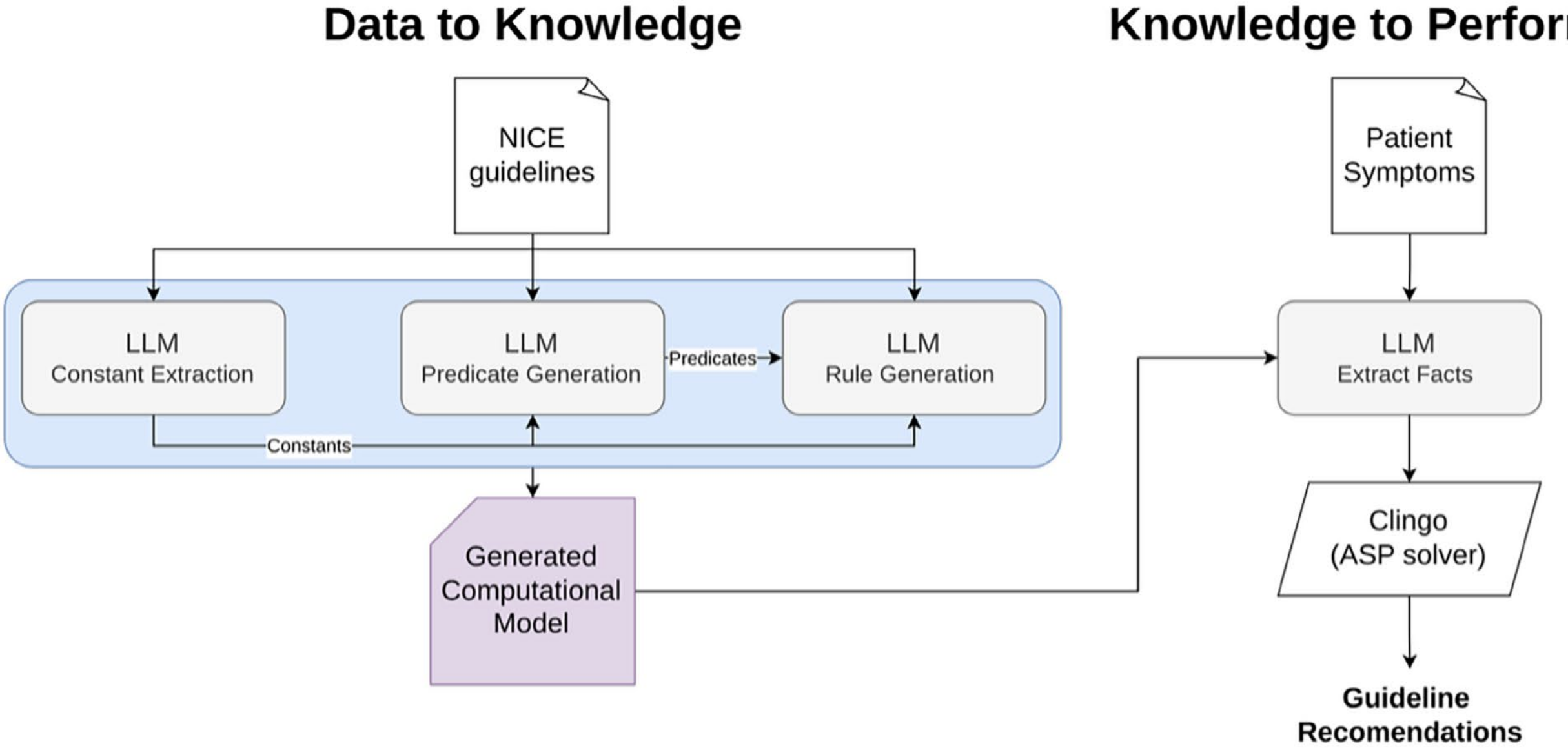

**FIGURE 1** | Approach for automatically converting clinical guidelines to an executable model. The data-to-knowledge approach uses the NICE guidelines as input and extracts constants, generates predicates and then generates rules in a stepwise manner to build a patient agnostic executable model. The knowledge-to-performance approach then takes free-form patient vignettes from which it extracts facts and combines them with the executable model to compute explainable guideline recommendations by means of a reasoning engine (the Clingo ASP solver) [15].

**Constant Extraction - In-context example**

Problem 1:
People without jaundice who have pancreatic abnormalities on imaging, offer a pancreatic protocol CT scan to people with pancreatic abnormalities but no jaundice. If the diagnosis is still unclear offer FDG-PET/CT and/or EUS with EUS-guided tissue sampling.

Constants:

- Imaging: pancreatic protocol CT
- Finding of the Pancreas: pancreatic abnormalities
- Symptoms: jaundice
- Procedures: FDG-PET/CT; EUS; EUS with tissue
- General Findings: unclear diagnosis

**FIGURE 2** | An in-context example used in the constant extraction step. Each in-context example includes a rule from the NICE guidelines and the corresponding categories and associated constants.

range of ASP rule types required for modeling clinical guidelines. Second, prompts were structured to maximize LLM reliability through explicit task decomposition, lightweight reasoning explanations, and warnings about common failure modes.

## 2.3 | Constant Extraction

The first step extracts domain-specific constants, such as diseases and symptoms, from the guideline text. The LLM is prompted with the NICE guideline text alongside a predefined set of constant categories, each accompanied by a brief description. These categories are based on a lightweight ontology aligned with SNOMED-CT concepts and the step ensures that all clinically relevant entities are explicitly found before any relational structure is generated.

## 2.4 | Predicate Generation

The second step identifies predicates that express relationships over the extracted constants. Using the same NICE guideline text and the constants identified in the first step, the LLM is prompted to define predicate schemas that capture how entities interact within the NICE guideline logic (e.g., symptoms possessed by a patient, procedures offered, findings derived from imaging). This step results in a reusable vocabulary of relations that form the backbone of the executable model.

Predicate Generation - In-context example

Problem 1:
People without jaundice who have pancreatic abnormalities on imaging, offer a pancreatic protocol CT scan to people with pancreatic abnormalities but no jaundice. If the diagnosis is still unclear offer FDG-PET/CT and/or EUS with EUS-guided tissue sampling.
Constants:
<CONSTANTS>
The categories in Constants include imaging, finding of pancreas, symptoms, procedures and general findings. We use different variables I, Fp, S, P and Gf to represent imaging, finding of pancreas, symptoms, procedures and general findings.
From the given problem, we offer imaging I or procedure P, have a symptom S or finding of pancreas Fp, findings Gf from imaging I, and offered imaging I.

Predicates:

- offer(I)
- offer(P)
- have(Fp)
- have(S)
- findings_from_imaging(I, Gf)
- offered(I)

**FIGURE 3** | An in-context example used in the predicate generation step. Each example includes a rule from the NICE guidelines, the constants extracted in the first step and the corresponding predicates.

Rule Generation - In-context example

Problem 1:
People without jaundice who have pancreatic abnormalities on imaging, offer a pancreatic protocol CT scan to people with pancreatic abnormalities but no jaundice. If the diagnosis is still unclear offer FDG-PET/CT and/or EUS with EUS-guided tissue sampling.
Constants: <CONSTANTS>
Predicates: <PREDICATES>
Given the Constants and the Predicates, the ASP rule for this statement is:

```
offer("pancreatic_protocol_ct"):-
    have("pancreatic_abnormalities"),
    not have("jaundice").

1{{offer("FDG-PET/CT"); offer("EUS_with_tissue")}}2:-
    have("pancreatic_abnormalities"),
    not have("jaundice"),
    offered("pancreatic_protocol_ct"),
    findings_from_procedure("pancreatic_protocol_ct", "diagnosis_unclear").
```

**FIGURE 4** | An in-context example used in rule generation step. Each example includes a rule from the NICE guidelines, the constants extracted in the first step and the predicates generated in the second step, together with the ASP rule representation of the chosen NICE guideline.

## 2.5 | Rule Generation

In the third and final step of D2K, the LLM generates ASP rules that formalize the conditional logic expressed in the NICE guidelines. The prompt provides the guideline text, the extracted constants, the generated predicates, and instructs the LLM model to encode the clinical recommendations as ASP rules.

## 2.6 | Knowledge to Performance

The output of the D2K stage is a patient-agnostic executable model, expressed in ASP, and encoding the NICE guideline logic. To enable patient-specific recommendations, the K2P stage translates unstructured patient record (in our approach this is expressed as a vignette) into a set of structured patient facts, expressed in the same formalism as that used to represent the generated executable model. This is achieved by prompting the LLM with both the generated executable model and the text description of the patient record, and instructing it to generate only those facts that correspond to predicates appearing in the body of the rules of the executable model, producing a set of atoms expressed using the same constants and predicates as the NICE guideline model. K2P combines these generated patient facts with the generated executable model and computes recommendations by means of a reasoning engine, the ASP solver Clingo [15], yielding patient-specific recommendations that can be directly traced back to individual guideline rules.

## 2.7 | Experiments and Evaluations

### 2.7.1 | D2K

We applied the D2K approach on the NICE guidelines for both pancreatic cancer (PC) and lung cancer (LC). We have used Claude Opus 4.5, the most capable instruction-following LLM model available at the time of the study. In-context examples were derived exclusively from the PC NICE guidelines and reused unchanged for LC enabling an assessment of cross-guideline generalization. Two ablation studies were conducted: zero-shot generation, in which no in-context examples were provided, and direct in-context generation, in which the LLM generated ASP rules from the NICE guidelines in a single step using the same examples but without the structured multi-stage D2K approach.

ASP encodings may yield equivalent stable models, meaning there is not one gold standard of translation. The objective of the evaluation is to ensure that each generated rule faithfully captures the clinical condition–recommendation relationship expressed in the source guideline. Accordingly, each generated rule was independently evaluated by three clinicians. Reviewers assessed whether the rule body correctly represented the clinical conditions described in the guideline and whether the rule head reflected the intended recommendation. Inter-reviewer agreement was quantified using Cohen's $\kappa$ coefficient. To support this assessment, clinicians were provided with an introduction to ASP rule syntax and were shown the original guideline statement alongside the corresponding ASP rule and a controlled NL rendering of the rule (Figure 5). Because the generated rules are simple conditional encodings of individual guideline recommendations and do not involve complex ASP constructs, their interpretation aligns closely with the underlying clinical "if–then" logic. This representation was therefore sufficient for clinicians to assess the fidelity of each rule to the original guideline statement. Each rule was then categorized into one of five categories:

1. *No translation*—NICE guideline statement did not lead to a rule.
2. *Correct translation*—completely correct translation of the NICE guideline.
3. *Incorrect translation*—medically or logically incorrect translation.
4. *Missing key information*—incomplete translation but still clinically effective.
5. *Hallucinated content*—text mentioned in the generated rule was not present in the guideline.

Reviewers were also able to provide free-text comments to highlight issues not captured by these categories. This evaluation protocol aligns with the fidelity assessment requirements described by Scott et al. [6].

**Original NICE Guideline Rule**

Consider an urgent CT scan in people aged 60 and over with weight loss and any of the following:

• diarrhoea • back pain • abdominal pain • nausea.
• vomiting • constipation • new-onset diabetes.

**Generated ASP Rule**

```
consider("CT", "urgent") :-
        age(X), X>=60, have("weight loss"),
        1{
                have("diarrhoea"); have("back pain");
                have("abdominal pain"); have("nausea");
                have("vomiting"); have("constipation");
                have("new-onset diabetes")
        }7.
```

**Controlled Natural Language of Generated Rule**

Relevant Patient Features
age >= 60 and have("weight loss") and at least 1 and at most 7 of: have("diarrhoea"), have("back pain"), have("abdominal pain"), have("nausea"), have("vomiting"), have("constipation"), have("new-onset diabetes"),

Action
consider("CT", "urgent").

**FIGURE 5** | During evaluation, the NICE guideline rules, generated ASP rules, and controlled natural language of the generated rule are provided to the reviewers to categorize the ASP translation of the original rule. This same process can be applied again during Knowledge-to-Performance for the generated recommendations.

#### 2.7.2 | K2P

To evaluate the K2P stage, we combined the generated executable model of the NICE guidelines with 20 PC patient vignettes and used a reasoning engine to infer patient-specific recommendations. The inferred outputs were compared against recommendations obtained from clinician-validated ground-truth interpretation of the guidelines. Recommendations inferred by both the ASP model and the ground truth were considered true positives. Recommendations inferred only by the generated model were treated as false positives, while recommendations present only in the ground truth were treated as false negatives. Because true negatives are not well-defined in this context, we report precision, recall, and $F1$ score rather than accuracy.

## 3 | Results

### 3.1 | D2K

The quality of the generated executable models is summarized in Figure 6. The PC guidelines were translated into 40 rules using the D2K pipeline and 42 rules using the direct in-context approach, while the LC guidelines resulted in 56 and 59 rules, respectively. The small differences in rule counts arise because the D2K pipeline occasionally produces more compact rules using ASP choice constructs, whereas the direct in-context baseline tends to split equivalent logic into multiple separate rules. Across both guidelines, the proposed D2K pipeline outperformed the direct in-context baseline in the human expert evaluation. The zero-shot baseline was excluded from this evaluation because it failed to produce syntactically valid ASP programs. Reviewer agreement under Cohen's $\kappa$ coefficient was moderate to high [20], Table 1.

Overall performance was lower for the LC guidelines, reflecting the greater diversity and heterogeneity of rule structures in these NICE guidelines. For LC, the in-context baseline achieved a slightly higher overall accuracy than the D2K approach. A Welch's $t$-test revealed no statistically significant difference between the D2K and in-context results, Appendix. Although comparable results, D2K resulted in a notably lower proportion of untranslated cases. While this did not aid correctness, it indicates that D2K attempted to formalize a broader set of guideline recommendations. This behavior is desirable in practice, as partially incorrect rules are typically easier to inspect and correct than entirely missing ones. Based on evaluator feedback, we further analyzed the translation errors produced by D2K which is presented in the Discussion.

### 3.2 | K2P

The 20 PC patient vignettes demonstrated a high degree of similarity between the generated rules and the ground truth rules used that generate recommendations, as detailed in Table 2, with patient specific metrics shown in Appendix.

## 4 | Discussion

### 4.1 | Weaknesses

The most common error category was the omission of a key component within an otherwise plausible rule. Typical examples included insufficiently specific predicates, such as generating *need*("surveillance") rather than *need*("surveillance," "pancreatic cancer"), or omitting qualifying adjectives such as "urgent," reflecting a limited flexibility in varying predicate arity. Higher-arity predicates (e.g., *need/2*) were rarely inferred, constraining the ability to encode richer contextual information within rules. This limitation also helps explain the tendency to preserve semantically dense phrases as long-form constants rather than decomposing them into more structured representations. For example, the phrase "enhancing solid component in the cyst" was retained verbatim, rather than being abstracted into a more compositional form such as *have*("pancreatic cyst," "solid component"). Expanding or refining constant categories and introducing in-context examples with hierarchical representations may help assist in these issues.

Encouragingly, hallucinations were rare. One likely explanation is that the stepwise structure of the D2K pipeline constrains the LLM to operate within previously generated constants and predicates, reducing opportunities for unsupported generation. The single observed hallucination for the PC guidelines involved generating "pancreatic protocol CT" in place of a generic "CT," likely because the former appeared frequently in nearby guideline text and was overgeneralised.

The approach struggled with rules involving multiple temporal or procedural components. For example, in recommendations such as "offer a pancreatic protocol CT scan before draining the bile duct," the LLM consistently failed to represent bile duct drainage as a separate procedural action. These types of errors reflect the difficulty of translating implicit procedural relationships that are readily inferred by humans but not explicitly stated in the text.

A subset of guideline recommendations resulted in no translation. This was particularly evident for rules that diverged from a conditional "if–then" structure or represented general, system-level commentary, such as "Every cancer alliance should have a system of rapid access to PET-CT scanning." Such statements do not readily map to patient-level ASP rules and highlight an inherent limitation in representing all guideline content within a single formalism. The rate of untranslated rules was higher for the LC guidelines, which exhibit greater diversity in presentation and structure than the PC guidelines. Clinical guidelines frequently contain statements that do not directly correspond to individual patient decision-making and heterogeneous rule structures. Moreover, there are often multiple valid ways to encode the same guideline recommendation in ASP. Taken together, these factors make it unrealistic to expect a fully automated approach to produce a completely correct computational model without human oversight. Expert review and iterative refinement are therefore to remain necessary components of any safe deployment.

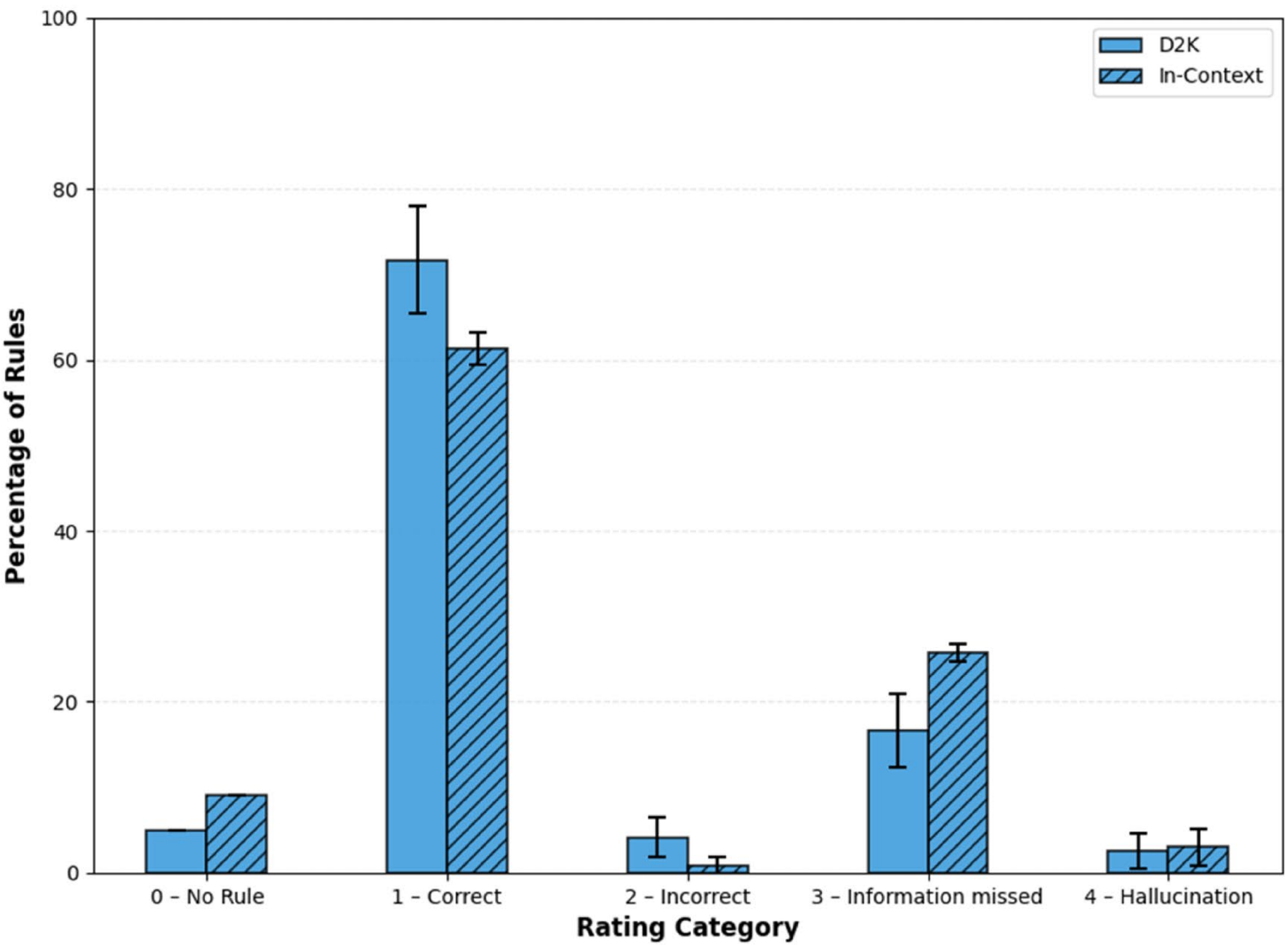

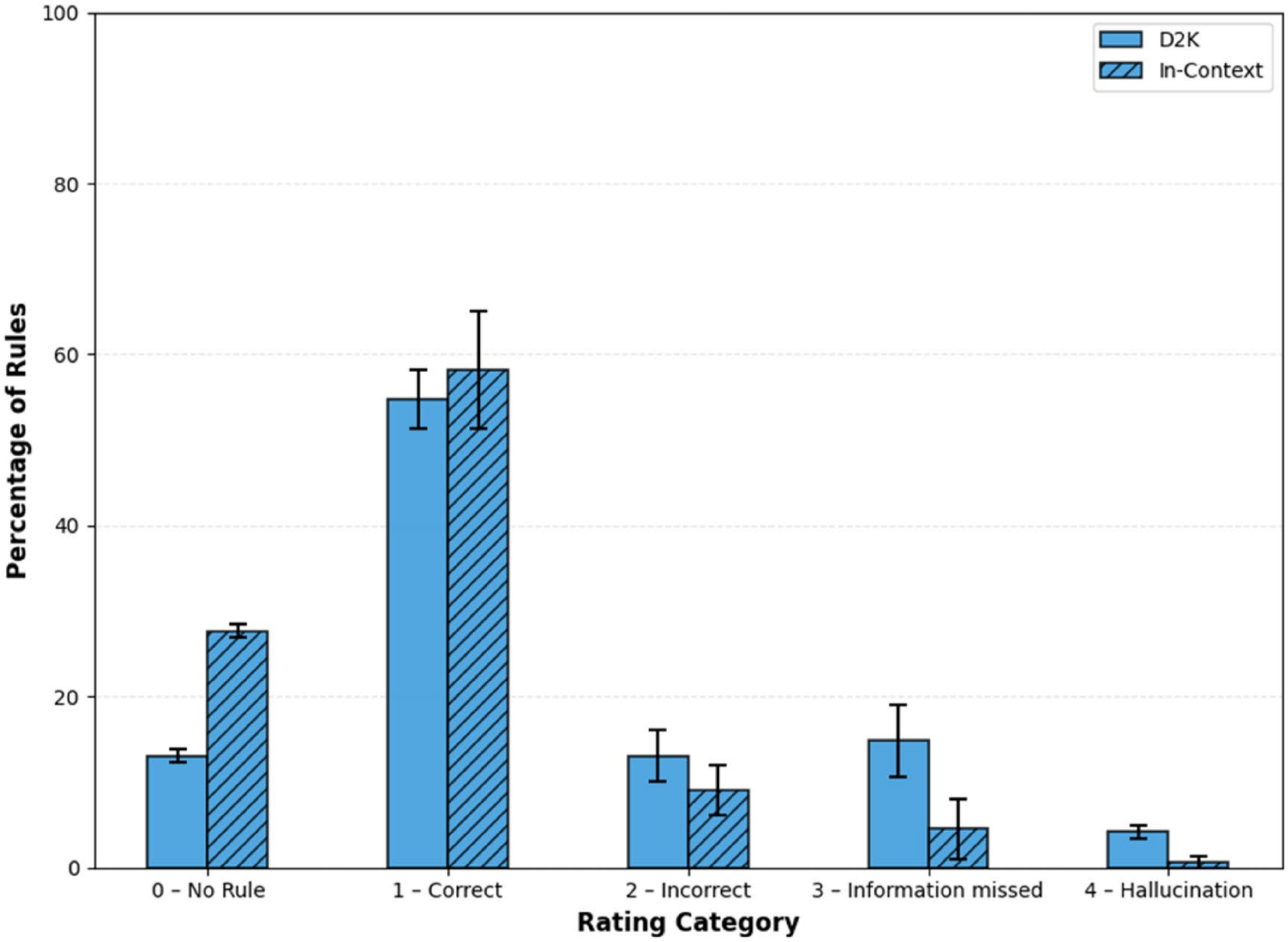

**FIGURE 6** | Human evaluation of generated executable models for the pancreatic cancer (top) and lung cancer NICE guidelines (bottom) for both the D2K stage and the in-context examples only approach. Expert reviewers were asked to categorize the generated rules into one of five categories: no rule, correct, incorrect, information missed, and hallucinated content.

**TABLE 1** | Cohen's $\kappa$ coefficient for the manual assessment of the in-context and D2K rules produced under both the pancreatic cancer and lung cancer settings by $n=3$ reviewers.

| | Cohen $\kappa$ coefficient | |
|---|---|---|
| | Pancreatic cancer | Lung cancer |
| In-context | 0.730 | 0.857 |
| D2K | 0.716 | 0.716 |

**TABLE 2** | Precision, recall, and $F1$ scores for the 20 pancreatic cancer patient vignettes.

| Metric | Average value |
|---|---|
| Precision | $0.894 \pm 0.115$ |
| Recall | $0.766 \pm 0.190$ |
| $F1$ score | $0.825 \pm 0.127$ |

In the K2P stage, alignment between the patient-agnostic guideline model and patient-specific data relied on LLM-based in-context learning. The observed errors, such as rules firing when they should not or failing to fire when expected, can mostly be attributed to missing or extraneous patient facts appended to the program rather than errors in the generated programs. Developing a more robust and systematic mechanism for extracting and validating patient features would likely improve downstream performance. Finally, while the patient vignettes were derived from real-world cases, they collectively activated only approximately 64% of the full guideline rule set. This limited coverage constrains the completeness of the evaluation but still provides the first evidence of practical utility.

### 4.2 | Strengths

A key strength of the proposed D2K approach is its modular, stepwise design, which enables targeted customisation and inspection at each stage of guideline translation. Each step of the approach can be modified or replaced with specialized or fine-tuned models for the specific task. Despite its relative simplicity, D2K demonstrated strong performance across two distinct guideline domains, highlighting its understanding of clinical guideline texts. As the approach relies exclusively on in-context learning, D2K can directly benefit from ongoing improvements in state-of-the-art LLMs without retraining, supporting long-term maintainability and portability. In addition, the D2K output is patient-agnostic, enabling reuse across cohorts.

Across both guideline sets, all generated rules were syntactically valid and logically coherent and produced stable models. With appropriately constructed in-context examples, the D2K approach successfully produced complex rules involving numerical thresholds, choice constructs, and familial relationships, suggesting that the prompting strategy effectively captured both the structure and semantics of guideline recommendations.

D2K is also highly amenable to iterative refinement. Because constants and predicates are explicitly extracted and stored as intermediate artifacts, domain experts can readily inspect and correct upstream abstractions before errors propagate into rule generation. Further to this transparency, every output recommendation can be traced back to the specific rule bodies that triggered it as demonstrated in Figure 5. This inherent explainability and adaptability in ASP is extremely valuable, as clinicians can audit the reasoning process and ensure that the rules are safe and recommendations are correct, increasing trust in the approach.

The D2K approach outperformed zero-shot and standard in context baselines. Furthermore, K2P evaluation demonstrated strong downstream utility, achieving an $F1$ score of 82.5% when applied to patient vignettes, which is promising for a first, fully automated system.

### 4.3 | Future Work and Implications

Building on these findings, future work should evaluate the proposed system using a broader and more diverse set of patient vignettes. While the current vignettes reflect common clinical trajectories, rare presentations and edge cases remain untested.

Beyond patient-level evaluation, the D2K approach has potential implications for guideline formalization itself. NICE has outlined a strategic vision for developing continuously learning, data-driven guideline systems [21]. One notable observation in this study is the performance gap between the PC and LC guidelines. A likely contributing factor is the difference in linguistic structure and presentation across guideline documents, which reduces the transferability of in-context examples. From a guideline engineering perspective, systematically applying the D2K approach across multiple guidelines could help identify vague, underspecified, or inconsistently phrased recommendations, as reflected by omitted or poorly translated rules. In this way, automated translation may also serve as a tool for guideline quality and clarity that could be deployed by NICE during guideline production.

A systematic comparison of the time and effort required for manual guideline curation versus the proposed automated approach would help quantify practical efficiency gains. This could be done to a previously unmodelled guideline to show the value of the D2K approach in development of computable clinical models.

The system is not yet ready for direct clinical integration. Further work is required to support clinician-led validation, improve the robustness of data-to-predicate alignment, and accommodate the diversity of guideline structures. A plausible deployment pathway would involve packaging the ASP-based guideline model as a digital knowledge object that can be executed within clinical decision support systems. In such a setting, structured patient features extracted from electronic health records (EHRs) derived patient features are appended to the guideline program to generate patient-specific recommendations. This integration also raises a limitation related to the closed-world assumption used in ASP reasoning, where facts not explicitly present in the program are treated as false through negation-as-failure. In real-world settings, EHR data may be incomplete, meaning that missing observations do not necessarily imply the absence of a clinical condition, consequently influencing which rules are triggered during inference. Practical deployments would therefore require clinician oversight to ensure that relevant clinical information is appropriately represented, including distinguishing between absent, unknown, and explicitly negated conditions, so that recommendations remain consistent with the full clinical context.

### Acknowledgments

We thank Professor Philip Crosbie and Professor Stephen Pereira for providing the patient vignettes used in this study. This work is supported by UK Research and Innovation (UKRI AI Centre for Doctoral Training in Digital Healthcare grant number EP/Y030974/1).

### Funding

This work was supported by UK Research and Innovation, EP/Y030974/1 and NIHR Imperial Biomedical Research Centre.

### Conflicts of Interest

The authors declare no conflicts of interest.

### Data Availability Statement

The NICE2ASP computational biomedical knowledge base is publicly available at https://github.com/Ashvin-Gupta/NICE-2-ASP which includes the prompts used, all results, 20 patient vignettes for K2P evaluation (no PHI included), ground truth ASP programs, and documentation. The test data does not contain any protected health information. Detailed information about the repository can be found in the GitHub repo itself and in Appendix.

## Appendix A

### Detailed Pipeline Execution and Data Documentation

#### Installation and Setup

The NICE2ASP2 pipeline is implemented in Python 3.12. Users can clone the repository from GitHub and are encouraged to run the pipeline using Docker, supported via the included Dockerfile and docker-compose.yml. Alternatively, users can set up a virtual environment, install dependencies via pip install -r requirements.txt, and install the Clingo ASP solver separately. Before running the pipeline, users must configure their API key by duplicating the .env.example file to create a .env file and inserting their valid ANTHROPIC_API_KEY. The repository includes comprehensive documentation, featuring a main README.md and additional README.md files in major subdirectories.

#### Input Data Structure

Input data is organized in the src/input_files/ directory. The input_guidelines/ subdirectory contains lung_cancer_guidelines.txt and pancreatic_cancer_guidelines.txt, which serve as the natural language source material for ASP generation. The D2K/ subdirectory houses five prompt templates that guide the LLM's behavior: constant_prompt.txt, predicate_prompt.txt, rule_generation_prompt.txt, in_context.txt, and zero_shot.txt. The ground_truths/ subdirectory contains human-authored ASP programs (GT_LC.lp and GT_PC.lp) serving as gold standards, alongside K2P_ground_truth.csv, which specifies the expected rule firings for patient vignettes. Finally, the K2P/ subdirectory contains PC_descriptions.txt (20 patient case descriptions) and setupProgram.txt.

#### Configuration and Execution

The approach uses separate YAML configuration files stored in src/configs/. lung_cancer_config.yaml is configured for "D2K-only" execution, while pancreatic_cancer_config.yaml supports D2K-only, K2P-only, and D2K+K2P modes. Each configuration file specifies the LLM model, pipeline version, and input paths. To execute the pipeline, reviewers run python main.py --config src/configs/lung_cancer_config.yaml (or the pancreatic equivalent). The pipeline provides verbose console output detailing progress through constant extraction, predicate extraction, rule generation, and patient vignette evaluation. A copy of the configuration file is automatically saved to the output directory for reproducibility.

#### Output Organization and Expected Results

All pipeline outputs are centralized in src/output_files/CLAUDE/. Evaluation metrics are stored in the reviews/ subdirectory, which contains:

- human review D2K/: Independent reviewer annotations with categorical scores and comments for generated rules.
- K2P review/: Per-patient prediction accuracy metrics saved as pancreatic cancer_K2P_review.csv.

Raw D2K outputs are organized by cancer type. The K2P/subdirectory (for pancreatic cancer) contains pipeline outputs including rulegen_response_fired.lp (ASP rules ready for Clingo), atoms.txt (patient facts), clingo_output.txt (solver results), and explanation.txt (natural language explanations).

#### Expected Output Formats and Content

The D2K pipeline outputs follow specific, verifiable formats. constant_response.txt contains categorized medical domain constants, predicate_response.txt lists typed predicate signatures and descriptions, and rulegen_response.txt contains a valid ASP program using standard syntax. The K2P pipeline outputs follow similar structured formats. rulegen_response_fired.lp appends tracking predicates to original rules, while atoms.txt contains patient-organized ASP facts. clingo_output.txt shows solver results for each patient, and explanation.txt provides clinical context detailing which rules fired and why.

#### Accessing Our Results

Reviewers can examine the generated ASP rules in CLAUDE/lung cancer/D2K/rulegen_response.txt and CLAUDE/pancreatic cancer/D2K/rulegen_response.txt. For K2P evaluation, per-patient prediction metrics are available in CLAUDE/reviews/K2P review/pancreatic cancer_K2P_review.csv. Detailed explanations of rule firings and complete Clingo solver outputs are found in CLAUDE/pancreatic cancer/K2P/explanation.txt and CLAUDE/pancreatic cancer/K2P/clingo_output.txt, respectively. Human reviewer annotations comparing the D2K pipeline and baseline methods are stored in CLAUDE/reviews/human review D2K/ across separate subdirectories for each independent reviewer.

#### Hypothesis Test

To evaluate whether the In-Context performance ($58.2 \pm 6.8\%$) is statistically higher than the D2K performance ($54.8 \pm 3.4\%$), a one-tailed Welch's $t$-test was conducted. This test was selected to account for the small sample size ($n = 3$ per group) and the inequality of variances between the two conditions.

The test was defined with a significance level of $\alpha = 0.05$.

- Group 1 (D2K): $\bar{x}_1 = 54.8$, $s_1 = 3.4$, $n_1 = 3$
- Group 2 (In-Context): $\bar{x}_2 = 58.2$, $s_2 = 6.8$, $n_2 = 3$
- Null Hypothesis ($H_0$): $\mu_2 \leq \mu_1$
- Alternative Hypothesis ($H_a$): $\mu_2 > \mu_1$

The standard error (SE) and t-score were calculated using the following formulas:

$$\text{Standard Error (SE)}: \text{SE} = \sqrt{\frac{s_1^2}{n_1} + \frac{s_2^2}{n_2}} = 4.389$$

$$t - \text{score}: t = \sqrt{\frac{(\bar{x}_2 - \bar{x}_1)}{\text{SE}}} = 0.7747$$

The degrees of freedom were calculated using the Welch-Satterthwaite approximation:

$$\text{d}f = \frac{\left(\frac{s_1^2}{n_1} + \frac{s_2^2}{n_2}\right)^2}{\frac{\left(\frac{s_1^2}{n_1}\right)^2}{n_1 - 1} + \frac{\left(\frac{s_2^2}{n_2}\right)^2}{n_2 - 1}} = 2.94$$

Using the $t$-distribution with $\text{d}f \approx 2.94$, the one-tailed p-value for $t = 0.775$ is: $p \approx 0.248$

Conclusion: Since $p > 0.05$, we fail to reject the null hypothesis. Despite the 3.4% increase in the mean for the In-Context group, the difference is not statistically significant given the high variance and the limited sample size.

#### Patient vignettes

**Vignette 1:** Patient is 63 year old male. Symptoms include abdominal pain, diabetes. On CT there were pancreatic abnormalities with a mass in the neck of the pancreas and the diagnosis unclear. On EUS there was desmoplastic ductal adenocarcinoma and on biopsy there was MMR wild-type.

**Vignette 2:** Patient is 54 year old female. Symptoms include abdominal pain. On CT there was liver metastases and pancreatic abnormalities with mass in the tail of the pancreas and the diagnosis was unclear. On EUS there was poorly differentiated adenocarcinoma.

**TABLE A1** | Predicted and ground truth rules that should fire for the 20 pancreatic cancer patient vignettes.

| Patient | TP | FP | FN | Precision | Recall | F1 | GT rules | Predicted rules | Missed rules | Extra rules |
|---|---|---|---|---|---|---|---|---|---|---|
| 1 | 2 | 0 | 0 | 1.00 | 1.00 | 1.00 | 1.1.18, 1.1.4 | 1.1.18, 1.1.4 | [] | [] |
| 2 | 5 | 1 | 2 | 0.83 | 0.71 | 0.77 | 1.1.18, 1.1.4, 1.1.5, 1.13.2, 1.13.2_B, 1.2.5 | 1.1.18, 1.1.4, 1.13.2, 1.13.2_B, 1.2.5, 1.2.5_B | 1.1.5, 1.3.3 | 1.2.5_B |
| 3 | 5 | 2 | 0 | 0.71 | 1.00 | 0.83 | 1.1.18, 1.1.7, 1.13.2, 1.13.2_B, 1.2.5 | 1.1.18, 1.1.4, 1.1.7, 1.13.2, 1.13.2_B, 1.2.5 | [] | 1.1.4, 1.2.5_B |
| 4 | 2 | 0 | 0 | 1.00 | 1.00 | 1.00 | 1.1.18, 1.2.4 | 1.1.18, 1.2.4 | [] | [] |
| 5 | 2 | 0 | 2 | 1.00 | 0.50 | 0.67 | 1.1.18, 1.1.4, 1.1.5, 1.1.6 | 1.1.18, 1.1.4 | 1.1.5, 1.1.6 | [] |
| 6 | 2 | 0 | 1 | 1.00 | 0.67 | 0.80 | 1.1.18, 1.1.4, 1.1.5 | 1.1.18, 1.1.4 | 1.1.5 | [] |
| 7 | 4 | 0 | 3 | 1.00 | 0.57 | 0.73 | 1.1.18, 1.1.4, 1.1.5, 1.1.6, 1.13.2, 1.13.2_B | 1.1.18, 1.13.2, 1.13.2_B, 1.2.4 | 1.1.4, 1.1.5, 1.1.6 | [] |
| 8 | 2 | 0 | 0 | 1.00 | 1.00 | 1.00 | 1.1.18, 1.1.4 | 1.1.18, 1.1.4 | [] | [] |
| 9 | 2 | 0 | 0 | 1.00 | 1.00 | 1.00 | 1.1.18, 1.1.4 | 1.1.18, 1.1.4 | [] | [] |
| 10 | 2 | 0 | 1 | 1.00 | 0.67 | 0.80 | 1.1.18, 1.1.4, 1.1.5 | 1.1.18, 1.1.4 | 1.1.5 | [] |
| 11 | 2 | 0 | 1 | 1.00 | 0.67 | 0.80 | 1.1.18, 1.1.4, 1.3.3 | 1.1.18, 1.1.4 | 1.3.3 | [] |
| 12 | 2 | 0 | 1 | 1.00 | 0.67 | 0.80 | 1.1.18, 1.1.4, 1.1.5 | 1.1.18, 1.1.4 | 1.1.5 | [] |
| 13 | 4 | 0 | 1 | 1.00 | 0.80 | 0.89 | 1.1.18, 1.1.4, 1.1.5, 1.13.2, 1.13.2_B | 1.1.18, 1.1.4, 1.13.2, 1.13.2_B | 1.1.5 | [] |
| 14 | 5 | 1 | 0 | 0.83 | 1.00 | 0.91 | 1.1.18, 1.1.4, 1.13.2, 1.13.2_B, 1.2.5 | 1.1.18, 1.1.4, 1.13.2, 1.13.2_B, 1.2.5, 1.2.5_B | [] | 1.2.5_B |
| 15 | 4 | 1 | 1 | 0.80 | 0.80 | 0.80 | 1.1.18, 1.1.4, 1.13.2, 1.13.2_B, 1.2.5 | 1.1.18, 1.13.2, 1.13.2_B, 1.2.5, 1.2.5_B | 1.1.4 | 1.2.5_B |
| 16 | 2 | 0 | 2 | 1.00 | 0.50 | 0.67 | 1.1.1, 1.1.18, 1.1.2, 1.1.3 | 1.1.18, 1.1.3 | 1.1.1, 1.1.2 | [] |
| 17 | 3 | 1 | 2 | 0.75 | 0.60 | 0.67 | 1.1.18, 1.1.7, 1.1.8_B, 1.1.9, 1.10 | 1.1.10, 1.1.18, 1.1.7, 1.1.8_B | 1.1.9, 1.10 | 1.1.10 |
| 18 | 4 | 0 | 0 | 1.00 | 1.00 | 1.00 | 1.1.18, 1.1.4, 1.13.2, 1.13.2_B | 1.1.18, 1.1.4, 1.13.2, 1.13.2_B | [] | [] |
| 19 | 3 | 0 | 0 | 1.00 | 1.00 | 1.00 | 1.1.18, 1.1.4, 1.3.3 | 1.1.18, 1.1.4, 1.3.3 | [] | [] |
| 20 | 2 | 1 | 1 | 0.67 | 0.67 | 0.67 | 1.1.18, 1.1.7, 1.1.8 | 1.1.18, 1.1.4, 1.1.7 | 1.1.8 | 1.1.4 |

**Vignette 3:** Patient is 77 year old male. Symptoms include obstructive jaundice. On CT there was biliary obstruction for which ECRP is being offered and the diagnosis was unclear. There is no tissue diagnosis and EUS showed a 40 mm mass and poorly differentiated adenocarcinoma.

**D2K Intermediate steps**

**Constant Extraction for Pancreatic Cancer**

Constants:

- Disease: pancreatic cancer, colorectal, gastro-oesophageal, lung, prostate, urological cancer, oesophageal, stomach, bladder, renal cancer.
- Symptoms: obstructive jaundice, jaundice, weight loss, appetite loss, diarrhoea, back pain, abdominal pain, nausea, vomiting, constipation, new-onset diabetes.
- Imaging: Imaging: pancreatic protocol CT, CT, ultrasound, MRI, MRI/MRCP.
- Procedures: FDG-PET/CT, EUS, EUS-guided tissue sampling, ERCP, biliary brushing, cytology, fine-needle aspiration, CEA assay, laparoscopy, laparoscopic ultrasound.
- Finding of pancreas: pancreatic abnormalities, pancreatic cysts, cystic lesions in the head of the pancreas, enhancing solid component in the cyst, main pancreatic duct that is 10 mm diameter or larger.

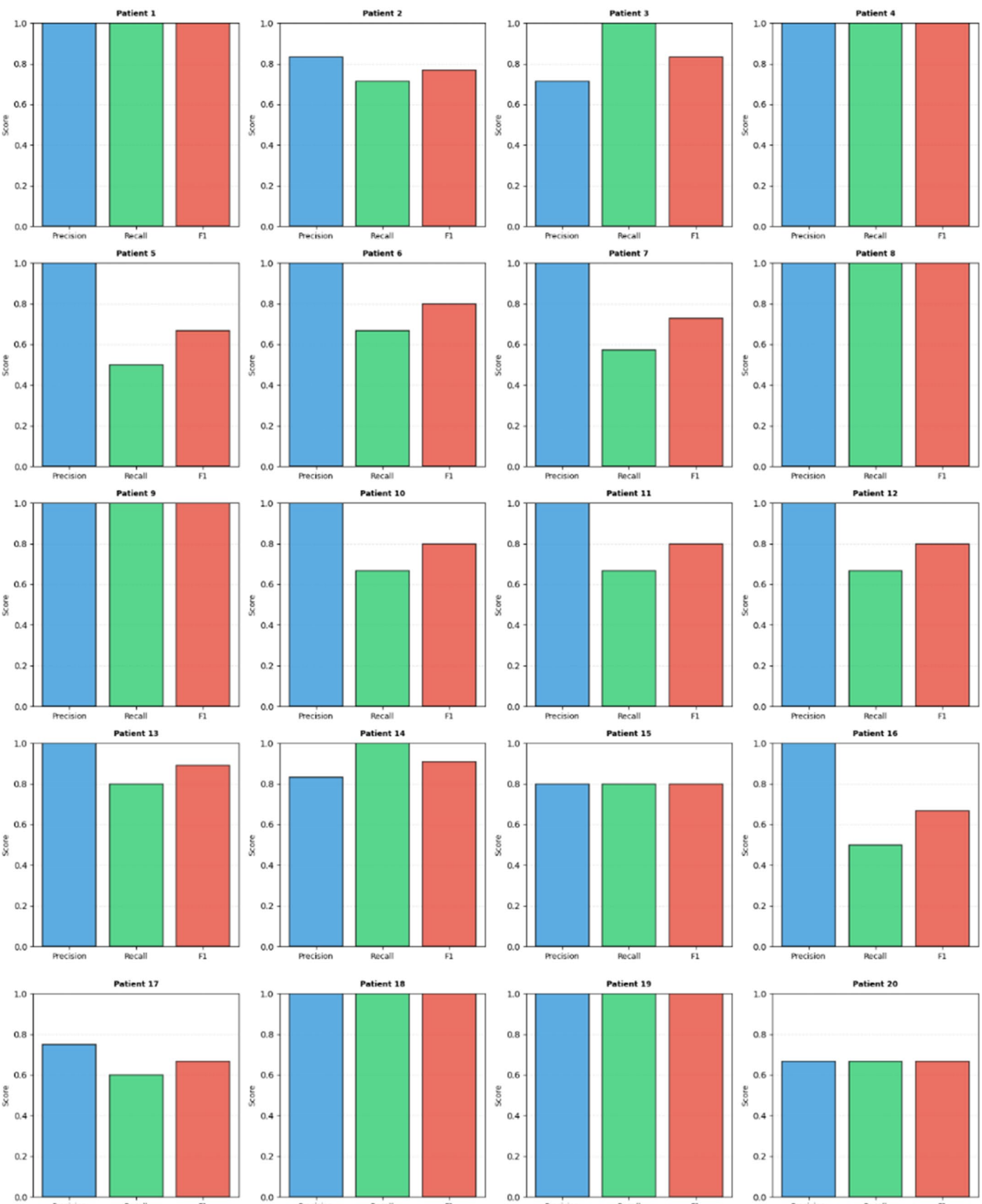

**FIGURE A1** | Average answer set performance metrics by patient vignette for pancreatic cancer guidelines between the ground truth answer sets and generated answer sets where metrics were defined as follows: true positives were atoms present in both the generated and ground truth answer sets; false positives were atoms present only in the generated set; and false negatives were atoms found only in the ground truth set.

- Genetic disease: hereditary pancreatitis, PRSS1, BRCA1, BRCA2, PALB2, CDKN2A (p16), Peutz–Jeghers syndrome, Lynch syndrome, MLH1, MSH2, MSH6, PMS2.
- Management: surveillance, suspected cancer pathway referral, urgent investigation.
- Treatments: surgery, radiotherapy, systemic therapy, resection, resectional surgery.
- Anatomical structures: bile duct, biliary obstruction, chest, abdomen, pelvis, liver metastases, peritoneal, tumour, node.
- Information about patient: 18, 40, 60, first-degree relatives, first-degree relatives with pancreatic cancer, inherited high risk, 2 or more first-degree relatives with pancreatic cancer, across 2 or more generations.
- General findings: unclear diagnosis, no tissue diagnosis, more information, more information on the likelihood of malignancy, localised disease, unexplained.

**Predicate Extraction for Pancreatic Cancer**

offer(I) offer(P) offer(P, P) offer(M, D) offer(Tr) offer(M) consider(I) consider(P) consider(P, P) consider(M, D) refer(M, D) refer(Tr) have(S) have(Fp) have(Gd) have(Ip) have(D) have(S, Gf) suspected(D) findings(I, Gf) findings(P, Gf) offered(I) offered(P) need(Gf) perform(P) drain(As) relieve(As) take(P, P) ask(Ip) assess(D)

**Prompts**

**Constant Prompt**

```
You are an expert in extracting and classifying medical constants from text.

Task plan:

1. Read the problem text carefully and identify all distinct medical concepts.
2. Classify each concept into one of the provided categories below.
3. Use the minimum exact wording from the text (do not paraphrase and no verbose phrases).
4. Output only the constants grouped by category in the specified format.

Category definitions:

1. Disease: Specific disease names (e.g., "pancreatic cancer").
2. Symptoms: Observable signs or symptoms (e.g., "jaundice", "weight loss").
3. Imaging: Diagnostic imaging modalities (e.g., "CT", "MRI").
4. Procedures: Medical interventions or diagnostic techniques (e.g., "EUS", "biopsy").
5. Finding_of_{ORGAN}: Imaging or pathology findings specific to {ORGAN}.
6. Genetic_disease: Inherited or genetic syndromes or mutations.
7. Management: Clinical management pathways or types of referral.
8. Treatments: Interventions for treating disease (e.g., "chemotherapy").
9. Anatomical_structures: Named body parts or regions.
10. Information_about_patient: Demographic or risk-related details (e.g., "40", "first-degree relatives").
11. General_findings: Concepts describing diagnostic uncertainty or test results (e.g., "unclear diagnosis", "more information").

Examples:

Example 1:

People without jaundice who have pancreatic abnormalities on imaging:

Offer a pancreatic protocol CT scan to people with pancreatic abnormalities but no jaundice. If the diagnosis is still unclear offer FDG-PET/CT and/or EUS with EUS-guided tissue sampling.

Constants:

1. Imaging: "pancreatic protocol CT".
2. Finding_of_pancreas: "pancreatic abnormalities".
3. Symptoms: "jaundice".
4. Procedures: "FDG-PET/CT", "EUS", "EUS-guided tissue sampling".
5. General_findings: "unclear diagnosis".

(Reasoning: "jaundice" → Symptom; "pancreatic protocol CT" → Imaging; "FDG-PET/CT" → Procedure; "unclear diagnosis" → General finding)

Example 2:

People with pancreatic cysts:

Consider fine-needle aspiration during EUS if more information on the likelihood of malignancy is needed.

Constants:

1. Procedures: "EUS", "fine-needle aspiration".
2. Understanding_of_diagnosis: "more information on the likelihood of malignancy".
3. Symptoms: "pancreatic cysts".

Example 3:

People with inherited high risk of pancreatic cancer:

Offer surveillance for pancreatic cancer to people with:

• hereditary pancreatitis and a PRSS1 mutation
• BRCA1 or BRCA2, and one or more firstdegree relatives with pancreatic cancer

Constants:

1. Disease: "pancreatic cancer".
2. Management: "surveillance".
3. Genetic_disease: "hereditary pancreatitis", "PRSS1", "BRCA1", "BRCA2".
4. Information_about_patient: "firstdegree relatives with pancreatic cancer", "inherited high risk".

Example 4:

People with inherited high risk of pancreatic cancer:
```

Do not offer EUS to detect pancreatic cancer in people with hereditary pancreatitis.

Constants:

1. Procedures: "EUS".
2. Genetic_disease: "hereditary pancreatitis".
3. Disease: "pancreatic cancer".
4. Information_about_patient: "inherited high risk".

Example 5:

Refer people using a suspected cancer pathway referral for pancreatic cancer if they are aged 40 and over and have jaundice.

Constants:

1. Disease: "pancreatic cancer".
2. Symptoms: "jaundice".
3. Information_about_patient: "40".
4. Management: "suspected cancer pathway referral".

Example 6:

Consider an urgent, direct access CT scan (to be done within 2 weeks), or an urgent ultrasound scan if CT is not available, to assess for pancreatic cancer in people aged 60 and over with weight loss and any of the following:

- diarrhoea
- back pain
- abdominal pain
- nausea
- vomiting
- constipation
- new-onset diabetes.

Constants:

1. Symptoms: "weight loss", "diarrhoea", "back pain", "abdominal pain", "nausea", "vomiting", "constipation", "new-onset diabetes".
2. Imaging: "CT", "ultrasound".
3. Information_about_patient: "60".
4. General_findings: "urgent".

Output Format:

Constants:

Category1: "constant1", "constant2".

Category2: "constant1", "constant2".

Problem to solve:

{problem_text}

#### Predicate Prompt

You are an expert in clinical logic and Answer Set Programming (ASP).

Task plan:

Your goal is to generate all necessary predicates that express relationships among the extracted constants from a clinical guideline statements.

1. Read the problem text carefully.
2. Review the provided constants and their categories.
3. Identify all actions, relationships, or conditions that connect constants.
4. Convert these relationships into predicates of the form predicate(X1, X2, ..., Xn) where each argument corresponds to a category (e.g., Imaging, Symptoms, Procedures, etc.).
5. Use the minimum number of predicates needed to capture all relationships (avoid redundancy).
6. Use the patterns and examples provided to ensure consistency.

General Guidance

1. Predicates typically correspond to verbs or key relationships in the guideline text.
2. In clinical guidelines, offer usually denotes a recommended action; have denotes patient features; findings encodes diagnostic results or follow-up conditions.
3. Temporal relationships (e.g., "if diagnosis still unclear after CT") should use the predicate offered(X) to indicate a previously performed procedure.
4. There should never be the word 'not' included in the predicate
5. Sometimes certain constants will have adjectives such as new-onset or unexplained or suspected. In this case can increase the ariof the predicate to two and have the rule as for example have(<disease>, <adjective>).

Variable Naming Convention

Use the following variable conventions throughout:

D – Disease

S – Symptoms

I – Imaging

P – Procedures

Fl – Finding_of_{ORGAN}

Gd – Genetic_disease

M – Management

Tr – Treatments

As – Anatomical_structures

Ip – Information_about_patient

Gf – General_findings

Examples:

Example 1:

People without jaundice who have pancreatic abnormalities on imaging:

Offer a pancreatic protocol CT scan to people with pancreatic abnormalities but no jaundice. If the diagnosis is still unclear offer FDG-PET/CT and/or EUS with EUS-guided tissue sampling.

Constants:

1. Imaging: "pancreatic protocol CT".

2. Finding _ of _ pancreas: "pancreatic abnormalities".
3. Symptoms: "jaundice".
4. Procedures: "FDG-PET/CT", "EUS", "EUS-guided tissue sampling".
5. General _ findings: "unclear diagnosis".

Predicates:

1. offer(I)
2. offer(P)
3. offer(P, P)
4. have(Fp)
5. have(S)
6. findings(I, Gf)
7. offered(I)

% Note the offered(I) predicate is VERY important here as it allows us to have temporal relationships within our rules.

% Note the predicate findings(I, Gf). This predicate should be used in situations where we have a finding Gf from an imaging I where the finding could be information such as 'diagnosis unclear', 'localised disease', 'malignant' etc.

% Note we can have multiple predicates with the same name but different arity such as offer(P), offer(P, P), where a higher arity predicate is able to express more specific relations.

Example 2:

People with pancreatic cysts:

Consider fine-needle aspiration during EUS if more information on the likelihood of malignancy is needed.

Constants:

1. Procedures: "EUS", "fine-needle aspiration".
2. Understanding _ of _ diagnosis: "more information on the likelihood of malignancy".
3. Symptoms: "pancreatic cysts".

Predicates:

1. consider(P, T)
2. need(Gf)
3. have(S)

Example 3:

People with inherited high risk of pancreatic cancer:

Offer surveillance for pancreatic cancer to people with:

• hereditary pancreatitis and a PRSS1 mutation

• BRCA1 or BRCA2, and one or more firstdegree relatives with pancreatic cancer

Constants:

Constants:

1. Disease: "pancreatic cancer".
2. Management: "surveillance".
3. Genetic _ disease: "hereditary pancreatitis", "PRSS1", "BRCA1", "BRCA2".
4. Information _ about _ patient: "firstdegree relatives with pancreatic cancer", "inherited high risk".

Predicates:

1. offer(M, D)
2. have(Gd)
3. have(Ip, D)
4. first _ degree _ relative(D, PersonID)

Example 4:

People with inherited high risk of pancreatic cancer:

Do not offer EUS to detect pancreatic cancer in people with hereditary pancreatitis.

Constants:

1. Procedures: "EUS".
2. Genetic _ disease: "hereditary pancreatitis".
3. Disease: "pancreatic cancer".
4. Information _ about _ patient: "inherited high risk".

Predicates:

1. offer(P)
2. have(Gd)
3. have(Ip, D)

Example 5:

Refer people using a suspected cancer pathway referral for pancreatic cancer if they are aged 40 and over and have jaundice.

Constants:

1. Disease: "pancreatic cancer".
2. Symptoms: "jaundice".
3. Information _ about _ patient: "40".
4. Management: "suspected cancer pathway referral".

Predicates:

1. refer(M, D)
2. have(S)
3. have(Ip)

Example 6:

Consider an urgent, direct access CT scan (to be done within 2 weeks), or an urgent ultrasound scan if CT is not available, to assess for pancreatic cancer in people aged 60 and over with weight loss and any of the following:

• diarrhoea
• back pain
• abdominal pain
• nausea
• vomiting
• constipation
• new-onset diabetes.

Constants:

1. Symptoms: "weight loss", "diarrhoea", "back pain", "abdominal pain", "nausea", "vomiting", "constipation", "new-onset diabetes".
2. Imaging: "CT", "ultrasound".
3. Information_about_patient: "60".
4. General_findings: "urgent".

Predicates:

1. consider(I, Gf)
2. have(S)

Self-Verification Checklist

Before producing the final answer:

1. Each predicate must be grounded in a verb or explicit relationship from the text.
2. All variables must correspond to one of the defined constant categories.
3. Avoid redundant or semantically overlapping predicates.
4. Ensure predicates generalize correctly across multiple rules (avoid overspecification).
5. Output only the list of predicates — no extra commentary or explanation.

Problem to solve:

{problem_text}

{processed_constants}

Predicates:

**Rule Generation Prompt**

You are an expert in Answer Set Programming (ASP) logic and clinical guideline formalization.

Task Plan

1. Carefully read the clinical guideline text in the Problem text.

2. Review the provided constants and predicates (already extracted).

3. Identify each logical condition, exception, and action in the text.

4. Express each as an ASP rule following the pattern used in the examples (Problems 1–6).

5. For negated recommendations ("Do not offer..."), produce constraints (:- ...).

6. Preserve temporal or conditional sequencing using offered(X) when a later action depends on a previous offer.

7. Output all resulting rules grouped by their guideline reference number, using the format:

```
[rule_number]
rule1.
rule2.
...
```

Background knowledge:

1. offered(X) :- offer(X).

This encodes temporal relationships between offered procedures or imaging steps.

Make sure rules are not in the form offer(X) :- offered(X) as this implies that we can only offer X if has already been offered which is impossible.

2. Age is represented as age(X), X>=number.

3. The notation 1{{A;B}}2 means choose at least 1 and at most 2 options from the set. Can use this in both the head (before :-) or in the body (after :-) and can help with simplification.

4. The symbol :- at the start of the line denotes a constraint (e.g anything atoms in the body cannot be exist together).

5. Use the constants and predicates provided — do not invent new ones.

6. The rules must be ASP valid. One condition is that the rules must be grounded, meaning if a variable appears in the head, it must also appear in a postive, non-built-in predicate in the body. If it appears in quatation marks it is fine.

e.g offer(P) :- body1, procedure(P) is correct. The procedure(P) is needed otherwise it is not grounded.

offer("ct scan") :- have("pain") is also correct.

Reasoning Pattern

1. Identify who the rule applies to (conditions → body).

2. Identify what is being offered, referred, or done (action → head).

3. Identify exceptions or negations (not have(S) or :- ...).

4. Identify temporal links between rules (offered(X), findings(I, Gf)).

Examples:

Example 1:

People without jaundice who have pancreatic abnormalities on imaging:

Offer a pancreatic protocol CT scan to people with pancreatic abnormalities but no jaundice. If the diagnosis is still unclear offer FDG-PET/CT and/or EUS with EUS-guided tissue sampling.

Constants:

1. Imaging: "pancreatic protocol CT".
2. Finding_of_pancreas: "pancreatic abnormalities".
3. Symptoms: "jaundice".
4. Procedures: "FDG-PET/CT", "EUS", "EUS-guided tissue sampling".
5. General_findings: "unclear diagnosis".

Predicates:

1. offer(I)

2. offer(P)
3. have(Fp)
4. have(S)
5. findings(I, Gf)
6. offered(I)

ASP:

offer("pancreatic protocol CT") :- have("pancreatic abnormalities"), not have("jaundice").

1 {{offer("FDG-PET/CT"); offer("EUS", "EUS-guided tissue sampling")}} 2 :- have("pancreatic abnormalities"), not have("jaundice"), offered("pancreatic protocol CT"), findings("pancreatic protocol CT", "diagnosis unclear").

Reasoning:

Main action: offer CT → offer("pancreatic protocol CT") from first statement. offer FDG-PET/CT or EUS-guided tissue sampling → offer("FDG-PET/CT"), offer("EUS-guided tissue sampling") from second statement.

Conditions: have abnormalities, not have jaundice from the first statement. findings of CT procedure means an unclear from the second statement.

Temporal rule: if CT was offered and diagnosis still unclear → offer further procedures. Symptoms from first conditions carry on over as statements are linked.

% These rules read as:

% If the patient has pancreatic abnormalities but no jaundice then offer a pancreatic protocol CT scan.

% If the findings from the pancreatic protocol CT are unclear then choose at least 1 but no more than 2 of the options in the curly brackets

% Note the offered(I) predicate is VERY important here as it allows us to have temporal relationships within our rules.

Example 2:

People with pancreatic cysts:

Consider fine-needle aspiration during EUS if more information on the likelihood of malignancy is needed.

Constants:

1. Procedures: "EUS", "fine-needle aspiration".
2. Understanding_of_diagnosis: "more information on the likelihood of malignancy".
3. Symptoms: "pancreatic cysts".

Predicates:

1. consider(P, T)
2. need(Gf)
3. have(S)

ASP:

consider("EUS", "fine-needle aspiration") :- need("more information on the likelihood of malignancy"), have("pancreatic cysts").

% This rule reads as:

% If the patient has pancreatic cysts and we need more information on the liklihood of malignancy then consider EUS with fine-needle aspiration.

Example 3:

People with inherited high risk of pancreatic cancer:

Offer surveillance for pancreatic cancer to people with:

• hereditary pancreatitis and a PRSS1 mutation

• BRCA1 or BRCA2, and one or more firstdegree relatives with pancreatic cancer

Constants:

Constants:

1. Disease: "pancreatic cancer".
2. Management: "surveillance".
3. Genetic_disease: "hereditary pancreatitis", "PRSS1", "BRCA1", "BRCA2".
4. Information_about_patient: "firstdegree relatives with pancreatic cancer", "inherited high risk".

Predicates:

1. offer(M, D)
2. have(Gd)
3. have(Ip, D)
4. first_degree_relative(D, PersonID)

ASP:

offer("surveillance", "pancreatic cancer") :- have("inherited high risk", "pancreatic cancer"), have("hereditary_pancreatitis"), have("PRSS1").

offer("surveillance", "pancreatic cancer") :- have("inherited high risk", "pancreatic cancer"), 1{{have("BRCA1"); have("BRCA2"); have("PALB2"); have("CDKN2A")}}4, first_degree_relative("pancreatic cancer", _).

% These rules read as:

% If the patient has inherited high risk, hereditary pancreatitis and PRSS1 then offer surveillance for pancreatic cancer.

% If the patient has inherited high risk, any first degree relaticve with pancreatic cancer and has any one of the genes BRCA1, BRCA2, PALB2, CDKN2A, then offer surveillance for pancreatic cancer.

% Note that we represent how many first degree relatives have a disease in the format first_degree_relative(D,X), where D is the disease and X is person X.

% Both D and X can be left blank, _, if we do not know the disease or the person. If there are two people that have the same diesase we could exploit this syntax and write X!=Y to show that they are different people.

Example 4:

People with inherited high risk of pancreatic cancer:

Do not offer EUS to detect pancreatic cancer in people with hereditary pancreatitis.

Constants:

1. Procedures: "EUS".
2. Genetic_disease: "hereditary pancreatitis".
3. Disease: "pancreatic cancer".
4. Information_about_patient: "inherited high risk".

Predicates:

1. offer(P)
2. have(Gd)
3. have(Ip, D)

ASP:

:- have("inherited high risk"), offer("EUS"), have("hereditary pancreatitis").

% This rule reads as:

% Do not offer EUS if the patient has inherited high risk and hereditary pancreatitis

Example 5:

Refer people using a suspected cancer pathway referral for pancreatic cancer if they are aged 40 and over and have jaundice.

Constants:

1. Disease: "pancreatic cancer".
2. Symptoms: "jaundice".
3. Information_about_patient: "40".
4. Management: "suspected cancer pathway referral".

Predicates:

1. refer(M, D)
2. have(S)
3. have(Ip)

ASP:

refer("suspected cancer pathway", "pancreatic cancer") :- age(X), have("jaundice"), X>=40.

% These rules read as:

% If the patient is above the age of 40 and has jaundice then refer to the suspected cancer pathway

% Note that we represent the age of the patient in the format age(X), where X is the age of the patient.

Example 6:

Consider an urgent, direct access CT scan (to be done within 2 weeks), or an urgent ultrasound scan if CT is not available, to assess for pancreatic cancer in people aged 60 and over with weight loss and any of the following:

- diarrhoea
- back pain
- abdominal pain
- nausea
- vomiting
- constipation
- new-onset diabetes.

Constants:

Constants:

1. Symptoms: "weight loss", "diarrhoea", "back pain", "abdominal pain", "nausea", "vomiting", "constipation", "new-onset diabetes".
2. Imaging: "CT", "ultrasound".
3. Information_about_patient: "60".
4. General_findings: "urgent".

Predicates:

1. consider(I, Gf)
2. have(S)

ASP:

consider("CT scan", "urgent") :- age(X), have("unexplained weight loss"), 1{{have("diarrhoea"); have("back pain"); have("abdominal pain"); have("nausea"); have("vomiting"); have("constipation"); have("new onset diabetes")}}7, X>60.

consider("ultrasound", "urgent") :- age(X), have("unexplained_weight_loss"), 1{{have("diarrhoea"); have("back pain"); have("abdominal pain"); have("nausea"); have("vomiting"); have("constipation"); have("new onset diabetes")}}7, X>60, not available("CT_scan").

% These rules are read as

% If the patient is over the age of 60, has unexplained weight loss and any one of diarrhoea, back pain, abdominal pain, nausea, vomiting, constipation or new-onset diabetes then consider an urgent CT scan.

% If the same conditions are present but the CT scan is not available then consider and ultrasound instead.

Problem to solve:

{problem_text}

{constants}

Predicates:

{predicates}